%% file: main.tex
\documentclass[manuscript,nonacm]{acmart}
\usepackage{times}
\usepackage{soul}
\usepackage{url}
\usepackage{booktabs}
\usepackage{algorithm}
\usepackage{algorithmic}
\begin{document}

\title{Trustworthy Human Computation: A Survey}

\author{Hisashi Kashima}
\authornote{All authors contributed equally to this survey.}
\email{kashima@i.kyoto-u.ac.jp}
\affiliation{%
  \institution{Kyoto University}
  \streetaddress{Yoshida Honmachi}
  \city{Kyoto}
  \state{Kyoto}
  \country{Japan}
  \postcode{606-8501}
}

\author{Satoshi Oyama}
\authornotemark[1]
\affiliation{%
  \institution{Hokkaido University}
  \streetaddress{Kita 14, Nishi 9, Kita-ku}
  \city{Sapporo}
  \state{Hokkaido}
  \country{Japan}
 }
\email{oyama@ist.hokudai.ac.jp}

\author{Hiromi Arai}
\authornotemark[1]
\affiliation{%
  \institution{RIKEN}
  \streetaddress{Nihonbashi 1-4-1}
  \city{Chuo-ku}
  \state{Tokyo} 
  \country{Japan}
  \postcode{103-0027}
  }
\email{hiromi.arai@riken.jp}

\author{Junichiro Mori}
\authornotemark[1]
\affiliation{%
  \institution{The University of Tokyo}
  \streetaddress{7-3-1 Hongo, Bunkyo-ku}
  \city{Tokyo}
  \country{Japan}
  \postcode{113-8656}
  }
\email{mori@mi.u-tokyo.ac.jp}

\begin{abstract}
Human computation is an approach to solving problems that prove difficult using AI only, and involves the cooperation of many humans. Because human computation requires close engagement with both ``human populations as users'' and ``human populations as driving forces,'' establishing mutual trust between AI and humans is an important issue to further the development of human computation. This survey lays the groundwork for the realization of trustworthy human computation. First, the trustworthiness of human computation as computing systems, that is, trust offered by humans to AI, is examined using the RAS (Reliability, Availability, and Serviceability) analogy, which define measures of trustworthiness in conventional computer systems. Next, the social trustworthiness provided by human computation systems to users or participants is discussed from the perspective of AI ethics, including fairness, privacy, and transparency. Then, we consider human--AI collaboration based on two-way trust, in which humans and AI build mutual trust and accomplish difficult tasks through reciprocal collaboration. Finally, future challenges and research directions for realizing trustworthy human computation are discussed.
\end{abstract}

\maketitle

\section{Trustworthiness in Human Computation}
\input{sec/introduction}

\section{Trustworthy Human Computation Systems}
\input{sec/RAS}

\section{Social Trust of Human Computation}
\input{sec/socialtrust}

\section{Collaborative Human Computation}
\input{sec/human-ai}

\section{Challenges and Open Questions}
\input{sec/challenges}

\section{Conclusion}
\input{sec/conclusion}

\clearpage
\bibliographystyle{abbrv}
\bibliography{reference}

\end{document}

%% file: sec/introduction.tex
\subsection*{Rise of Human Computation}
AI technologies centered on machine learning, as represented by the rise of deep learning, have made remarkable progress in recent years. Its application areas are not limited to information and communication related fields, including recommendation and web marketing, but have expanded across many disciplines to include manufacturing, transportation, medicine, and various science related fields where a variety of innovations are expected. In general, machine learning can only be applied to areas where system inputs and outputs can be formally described and where a large amount of data can be collected. Furthermore, similar to humans, AI decisions always contain errors. This indicates that human teaching, judgment, and feedback at important points are necessary in critical application domains, such as automated driving and medical diagnostics, where research and development is ongoing and human lives are at stake. This trend is more pronounced for complex and critical problems. Real-world problems often do not require AI solution only, but involve some form of human participation. In future, we will face more difficult real-world challenges that require a closer integration of AI and human-driven systems. Over the past decade, AI research has focused on enhancing machine intelligence, especially through deep learning. However, as the impact of AI on society grows, the interface between humans and AI is being discussed in terms of both technology and regulation. For example, Ethically Aligned Design~\cite{IEEE_EAD} proposed by IEEE,  Principles on Artificial Intelligence~\cite{OECD} by OECD, and Ethics Guidelines for Trustworthy AI~\cite{EU} by EU have been proposed to promote the use of AI with respect to humans.

The approach of enlisting the help of (often unspecified) large numbers of humans to solve problems that are difficult for AI to solve by itself is called ``human computation''~\cite{Law2011}. Early examples include ReCAPTCHA~\cite{ahn2008}, which solves the limitations of AI recognition performance with the help of humans through tasks hidden in an access control interface, and VizWiz~\cite{bigham2010vizwiz}, a human-in-the-loop visual question-answering application to aid the visually impaired. The development of human computation was strongly encouraged by the concept of crowdsourcing, which emerged around the same time as human computation, and general-purpose crowdsourcing platforms such as Amazon's Mechanical Turk, which facilitate on-demand employment of an unspecified number of crowd workers. In 2013, HCOMP (AAAI Conference on Human Computation and Crowdsourcing), an academic community dedicated to human computation, was established and steadily progressed.

Many of the early studies in the field of human computation involved testing ability of general public participation with standard capabilities to solve challenges that proved difficult for AI to solve, given the new crowdsourcing platforms. (Various early research studies have been summarized by Law and von Ahn ~\cite{Law2011}.) Subsequently, attempts were made to establish a systematic design theory for human-AI systems that went beyond a mere collection of successful and unsuccessful cases. The development of a general framework for building various human-computation systems~\cite{little2010exploring,little2010turkit,kittur2011crowdforge} is one example of such an attempt. (More details are provided in Chapter 2.)

\subsection*{Toward Trustworthy Human Computation}
Similar to the discussions on the reliability of AI, it is essential to discuss the trustworthiness of human computation to play a more active role in human society. Before discussing this issue, it is necessary to define trustworthiness in human computation, in which there are at least two kinds. One relates to computing systems, while the other is based on trust between human computation systems and humans who interact with them. The former makes human computation systems as versatile and trustworthy as conventional computer systems, while the latter is described as the social and ethical responsibilities of systems that interact with humans, such as fairness and accountability.

When we implement human computation, we presumably try to achieve a particular goal (perhaps one that is difficult to achieve by using conventional computer systems alone). To achieve this goal, it is desirable to handle human computation with the same trustworthiness as that of conventional computers. Such ``human computers'' will utilize the collective power of humans as its driving force. Because humans have a great deal of uncertainty compared with conventional computational units, controlling these uncertainties is directly related to the assurance of trustworthiness. In previous studies, such trustworthiness was discussed from various perspectives, such as the motivation to participate in human computation~\cite{mason2009financial,feyisetan2019beyond} and quality assurance of crowdsourced  results~\cite{snow2008cheap,whitehill2009}. 

Human computation involves not only ``human populations as driving forces'' but also ``human populations as users.'' The use of AI technology in various aspects of society is currently under consideration, which increases the demand for AI system trust in a social context. Human computation cannot be further expanded without the trust of the latter; conversely, this concept is not sustainable without the trust of the former.

Moreover, beyond the above mentioned trusts between human computation and humans from one side to the other, the ultimate goal of human computation is to solve more difficult problems by having mutual trust and collaborating with each other.
Collaboration between humans and AI is essential for truly trustworthy human computation, not to mention the anecdote of the victory of a mixed team in the freestyle chess games, in which humans and AI are freely paired up~\cite{brynjolfsson2011race}.

Based on the above considerations, this study will focus on the interactions with humans inside and outside of human computation systems, and organize various existing studies according to trustworthiness, thereby laying the groundwork for discussions aimed at trustworthy human computation.

\subsection*{Existing Surveys and Reviews}
Individual surveys have been conducted on human computation, crowdsourcing, and the relationship between humans and AI. This survey differs from previous studies in that it examines the trustworthiness of human computation and organizes the existing studies from this perspective.

An overview of early research on human computation can be found in the book by Law and von Ahn~\cite{Law2011} and the survey by Quinn and Bederson~\cite{quinn2011human}. The handbook~\cite{michelucci2013handbook} has collected a wide range of related topics. Daniel et al.~\cite{daniel2018quality} surveyed various methods for quality control of human computation.

Several survey papers summarized recent discussions on trustworthy AI. Kaur et al.~\cite{kaur2022trustworthy} conducted a comprehensive review of trustworthy AI, where they discussed the trustworthiness of AI from various perspectives, including ethical discussions such as fairness, human acceptability (explainability and human participation), and safety (privacy and security). Thiebes et al.~\cite{thiebes2021trustworthy} summarized the concepts and perspectives of a trustworthy AI. In addition, a survey on AI fairness (particularly in the context of machine learning) was conducted by Mehrabi et al.~\cite{10.1145/3457607}. Techniques to help humans understand the decisions of (often black-box) AI are discussed as XAI (explainable AI), and a survey on this topic has bee conducted by Guidott et al.~\cite{guidotti2018survey}.

Collaboration with humans is built on trust in reciprocal relationships between humans and AI. Surveys~\cite{vaughan2017making,mosqueira2022human} and  book~\cite{monarch2021human} on human-in-the-loop machine learning discuss human participation in the machine learning process. Taking this a step further, some surveys~\cite{seeber2020,lai2021towards,vereschak2021} were conducted on collaborative work and decision-making between AI and humans.

As mentioned above, there are various perspectives on the trustworthiness of AI; however, yet none of the above are based on the unique perspective of human computation.

\subsection*{Structure of This Survey}
\begin{figure}[t]
 \centering
  \includegraphics[width=120mm, trim=0mm 0mm 0mm 0mm]{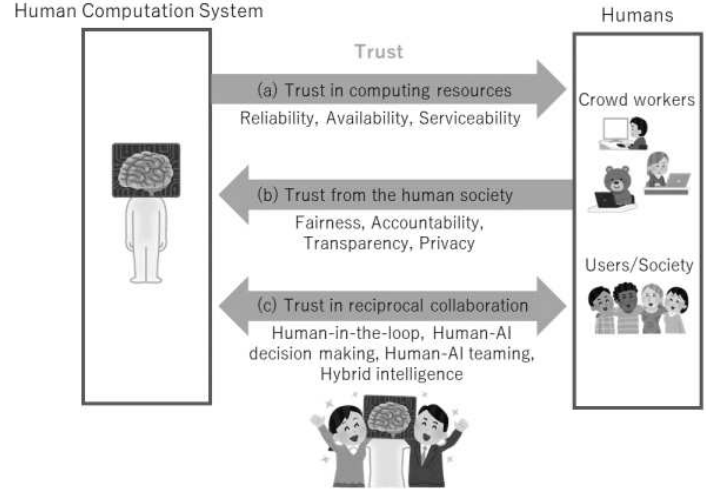}
  \caption{There are two types of trust in human computation: (a) trust from the human computation system to humans (crowd workers) as computational resources, and (b) trust from human groups (crowd workers, users, and society) to the human computation system.
  Based on these trusts, the goal of human computation is to solve difficult problems through (c) cooperation between both parties based on trust in reciprocal relationships.}
  \label{fig:concept}
\end{figure}

In the following three chapters, this survey summarizes the arguments for achieving trustworthy human computation from three perspectives: (a) trust in humans from the human computation system, (b) trust in human computation from humans, and (c) collaborative human computation based on trust in reciprocal relations between AIs and humans (Fig.\ref{fig:concept}).

In Chapter 2, we discuss the perspective of the trustworthiness of human computation as computing systems, that is, the trust possessed by humans in AI. In particular, we discuss human computation from the perspective of RAS (reliability, availability, and serviceability)~\cite{siewiorek1998reliable}, which are standard reliability and security measures in conventional computer systems.

Conversely, in Chapter 3, we discuss the trustworthiness that a human computation system offers to society and participants in human computation. With the social advancement of AI, there has been much discussion about the ethics AI and social trustworthiness of AI, such as fairness and privacy~\cite{kaur2022trustworthy}. This survey summarizes previous discussions, focusing on perspectives related to crowdsourcing workers who play an important role in the context of human computation.

In Chapter 4, we will discuss the collaborative intelligence perspective, in which humans and AI work together as a team to solve complex problems: first, human-in-the-loop human computation systems that explicitly treat human participation in AI systems; second, algorithm-in-the-loop (or decision supporting) human computation systems that focus on AI assisting humans; and finally, hybrid intelligence human computation systems that build mutually reciprocal relationships between AI and humans through collaboration. The discussions are then summarized relating to how trust should exist in collaborative human computation.

In Chapter 5, based on previous discussions, we present future challenges and research directions for achieving trustworthy human computation.

%% file: sec/RAS.tex
One aspect of human computation is the use of humans as a computing resource. In this chapter, we discuss the trustworthiness of human computation in computing systems driven by human labor, that is, in terms of the trustworthiness offered by humans to AI.

Although there have been many discussions on quality control in human computation~\cite{daniel2018quality}, metrics that consider multiple aspects of quality, as is the case in traditional computer systems, are not evident. One aspect of trustworthiness in human computation is the quality of a correct task execution. Human computation often uses crowdsourcing as a driving force. Crowdsourcing ranges from commercial, such as Amazon Mechanical Turk, to volunteer participation, such as citizen science projects. Many crowdsourcing tasks are designed such that they can be performed by workers who do not necessarily have expertise. However, the quality of tasks vary greatly depending on the knowledge and dedication of the workers. On commercial crowdsourcing platforms particularly, there are so-called ``spam workers'' who seek to maximize reward for as little effort and time as possible. To ensure quality in these situations, many studies have been conducted on how to select workers who produce high-quality deliverables and how to evaluate the quality of work. Furthermore, when considering the trustworthiness of human computation, it is necessary to consider not only the correctness of the computation results, but also a broader perspective. This includes the availability of results, or if there is a flaw in the computation, whether it can be corrected.

\begin{table}[t]
  \caption{Studies on reliability in human computation.}
  \label{table:ras-reliability}
  \centering
  \small
  \begin{tabular}{llr}
    \hline
    Category  & Reference  &  Description  \\
    \hline \hline
    Worker ability  & Kazai et al. (2011) \cite{kazai2011crowdsourcing} & Evaluating worker reliability by tasks with true answers\\
    & Ipeiriotis et al. (2010) \cite{Ipeiriotis2010b} & Identifying workers with malicious intent or misinterpretation of tasks\\
    & Raykar \& Yu (2011) \cite{Raykar2011} & Introducing the Spammer Score\\
    & Sheng et al. (2008) \cite{sheng2008get} & Conditions for successful majority voting\\
    & Snow et al. (2008) \cite{snow2008cheap} & Weighted majority voting by worker accuracy\\
      & Sakurai et al. (2013) \cite{sakurai13ability} & Discriminating accurate workers with different reward plans\\
    & Dawid \& Skene (1979) \cite{Dawid1979}  &  Estimating worker ability without true answers\\
    & Venanzi et al. (2014) \cite{Venanzi2014-fx} & Considering worker group structure\\
      & Li et al. (2019) \cite{Li2019-EBCC}  & Modeling correlations between workers\\
\hline
Cognitive bias & Draws et al. (2021) \cite{draws2021checklist} & Effects of cognitive bias\\
      & Newell \& Ruths (2016) \cite{newell2016} & Influence of the previous task\\
      & Eickhoff (2018) \cite{eickhoff2018} & Existence of various cognitive biases\\
      & Barbera et al. (2020) \cite{barbera2020} & Effect of confirmation bias\\
      & Coscia \& Rossi (2022) \cite{coscia2020} & Effect of confirmation bias\\
      & Demartini (2019) \cite{demartini2019} & Influence of worker attributes on fact-checking tasks\\
      & Biel \& Gatica-Perez (2014) \cite{biel2014} & Presence of bias in impression evaluation\\
      & Kulkarni et al.(2014) \cite{kulkarni2014} & Presence of bias in mutual evaluation\\
      & Eickhoff \& de Vries (2013) \cite{eickhoff2013} & Reducing cheat submissions\\
      & Hube et al. (2019) \cite{hube2019} & Making people aware of the presence of biases\\
      & Duan et al. (2020) \cite{duan2020does} & Working together with different perspectives\\
    & Faltings et al. (2014) \cite{faltings2014} & Influence of monetary incentives on bias\\
    & Gadiraju et al. (2017) \cite{gadiraju2017self} & Finding competent workers by the Dunning-Kruger effect\\
    & Gamalmaz \& Min (2021) & Eliminating the influence of confirmation bias\\
    & Echterhoff et al. (2022) \cite{echterhoff2022ai} & Eliminating the influence of anchoring bias\\
\hline
Task-worker relation      & Whitehill et al. (2009) \cite{whitehill2009} & Considering task difficulty\\
      & Welinder et al. (2010) \cite{welinder10muiti} & Considering multidimensional characteristics of tasks and workers\\ 
     & Bachrach et al. (2012) \cite{bachrach2012grade} & Bayesian model considering both worker ability and task difficulty\\ 
      & Oyama et al. (2013) \cite{oyama13accurate} & Considering workers' confidence about their answers\\
     & Ambati et al. (2011) \cite{Ambati2011} & Predicting  preferred tasks based on worker attributes\\
     & Yuen et al. (2012) \cite{yuen2012taskrec} & Recommendation using task browsing history and past tasks\\
    & Kulkarni et al. (2012) \cite{kulkarni2012mobileworks} & Recommendation of tasks by other workers\\
     & Li et al. (2014) \cite{wom} & Identifying appropriate workers for a task based on their attributes\\
& Bender \& Friedman (2018) \cite{bender2018data} & Different worker attributes between data collection and use\\    
& Sheng et al. (2008) \cite{sheng2008get} & Calculating the uncertainty of answers\\
& Donmez et al. (2009) \cite{Donmez2009} & Trade-off between exploration and exploitation of workers\\
\hline      
Various tasks      & Chen et al. (2013) \cite{chenWSDM2013} & Ranking aggregation from pairwise comparisons\\
      & Matsui et al. (2014) \cite{matsui14crowdordering} & Aggregation of multiple ranking lists\\
      & Wu et al. (2012) \cite{wu2012sembler} & Aggregation of multiple sequences \\
      & Wang \& Dang (2022) \cite{wang2022generative} & Aggregation of multiple text sentences \\
      & Baba \& Kashima (2013) \cite{baba13statistical} & Quality control using a generation-evaluation process\\
    \hline
  \end{tabular}
\end{table}

Human computation, which is a fundamental framework for humans and AI to cooperate in solving problems, is not able to compare and evaluate different systems from multiple perspectives because there is no standard quality evaluation metric like that for conventional computer systems. For human computation to be trusted and used in the real world as a more practical approach, it must be designed considering quality from various perspectives. In the field of computer engineering, RAS is used to evaluate systems from multiple perspectives to quantitatively and qualitatively assess the possibility of problems arising during system operations~\cite{siewiorek1998reliable}. However, previous human computation research has not comprehensively addressed the system trustworthiness from an RAS perspective. It is also unclear whether RAS, which targets systems consisting only of computers, can be directly applied to systems that include humans as components. From this point forward, we will discuss the differences between what should be considered in computer engineering and human computation for each aspect of RAS, and describe existing human computation studies that are relevant to each aspect. 

\subsection{Reliability}
In computer system design, reliability is defined by mean time to failure (MTTF), which is calculated by dividing the system uptime by the number of failures. However, hardware failures are rarely a problem in human computation, whereas human errors are far more frequent and of main concern. Therefore, the error rate of the task execution results would be a more appropriate criterion for reliability.

The research most related to the reliability in RAS is referred to as {\bf quality control} in human computation, and is commonly found in the literature. Statistical quality control, which improves the overall reliability through redundancy introduced by asking multiple workers to perform the same task, is the mainstay of quality control research~\cite{daniel2018quality}. Furthermore, an inherent problem with human computation is that the people participating are highly heterogeneous in terms of motivation. Therefore, research has also been conducted to design mechanisms that provide incentives for participants to work diligently from a game-theory perspective~\cite{muldoon2018survey}.

In human computation, if the results of some tasks are incorrect, it can be dismissed as long as the final result of the entire system's computation is correct. In the context of system design, this corresponds to a {\bf fault-tolerant} design in which individual component failures or operational defects do not significantly affect the overall process. Fault tolerance is a concept that is addressed in the field of reliability design. Quality control of human computation includes concepts pertaining to both (narrowly defined) quality control to enhance the quality of individual components and reliability design when multiple components are combined to build a system.

When controlling product quality at a manufacturing site, product samples are inspected for defects. Once defective products are found, the causes of them at the manufacturing site are investigated. For example, if defective products are concentrated in a particular production line, the line is stopped to investigate the cause. In crowdsourcing, which is a platform for human computation, people are the main cause task errors, and the work lines at the manufacturing site can be considered to correspond to crowdsourced workers.  (Of course, there can also be causes other than workers, such as unclear task descriptions or faulty work interfaces.) Therefore, quality modeling which focuses on workers has been widely used for crowdsourcing quality control. In the following subsections, we discuss existing research in terms of worker ability, cognitive bias, relationship between tasks and workers, and types of tasks. Table~\ref{table:ras-reliability} summarizes related studies based on the reliability of human computation.

\subsubsection{Considering worker abilities}
The most important factors in obtaining high-quality results is assigning tasks to workers with high ability. (Hereafter, we use the term ``ability'' not only in the narrow sense of worker knowledge and skills, but also in the broader sense that includes other factors that affect the accuracy of work results, such as worker morale.) In many crowdsourcing markets, workers are not paid until their work is reviewed and approved by the requester. If the requester is not satisfied with the results of the work, he/she may refuse to pay. The percentage of approved results submitted by a worker is an important indicator of the worker's ability to perform tasks. In many crowdsourcing markets, filtering functions are available, such as requesting workers with an approval rate of 95\

However, this method cannot be applied to new workers with a limited work history. In addition, if there is little relationship between previous and current tasks  (e.g., a task to answer a question in English and a task to answer a question in Chinese), the past evaluation may not be directly applicable to the estimation of ability to perform in the current task. 

If the correct answers to some of the tasks are known, they can be used to filter workers by evaluating their responses. Alternatively, worker abilities can often be measured by performing spot checks in which tasks with known answers are mixed with those with unknown answers~\cite{kazai2011crowdsourcing}. If a worker's ability is determined to be low, measures can be taken such as stopping further work or discarding results obtained from the worker.

Even if a worker performs a task by guesswork, the accuracy is not zero because there is a chance that he/she will obtain a correct answer by fluke. Therefore, workers whose answer rate is close to zero (i.e., workers who almost always answer the opposite of the correct answer) can be considered malicious workers or workers who misunderstand the question and provide incorrect answers. For example, in the task of classifying ``normal mail'' and ``junk mail,'' some workers mistakenly checked ``junk mail'' when they were instructed to check ``normal mail.'' However, if such workers can be identified using data with known correct answers, other correct answers can be estimated with high accuracy by replacing their answers in the opposite manner~\cite{Ipeiriotis2010b}.

On the other hand, workers who answer randomly may be so-called spam workers who answer without looking at the question because they only want money from the beginning, workers who answer by a guess without sufficient ability to answer, or bots or other programs other than humans. These workers answer randomly, independent of the question; thus, the probability of a worker answering the question does not depend on the correct answer. A spammer score that identifies such workers has been proposed~\cite{Raykar2011}.

The simplest way to provide robustness against errors in the results of worker tasks is to introduce redundancy by requesting the same task to multiple workers, rather than requesting it to only one worker, based on the idea that ``two heads are better than one.'' For example, in the case of a binary classification problem, if three people are requested to perform a task and a majority vote is taken, the correct answer can be obtained even if one person makes a mistake. It is important to note that the assumption for majority voting is that the percentage of correct answers by workers is high to some degree~\cite{sheng2008get}.

As described above, simple majority voting, in which a majority is determined by one person with one vote, can be regarded as assuming that all workers have equal ability (i.e., equal chance of obtaining the correct answers). However, worker abilities often vary with crowdsourcing. In such cases, it is a bad idea to treat the opinions of workers with a high and low probability of obtaining the correct answers equally when taking a majority vote. If the worker's accuracy is known from a pre-test using a task for which the correct answer is known, the possibility that the correct answer can be derived increases by assigning more weight to the opinion of the worker with higher accuracy. In fact, it has been shown that using such weighted majority voting in multiple natural language processing tasks can estimate the correct answer with higher accuracy than simple majority voting~\cite{snow2008cheap}.

For tasks that predict the future, the correctness/incorrectness of the worker's answers becomes known in the future. In such tasks, a method to select high-quality workers has been proposed by preparing two types of reward plans: a high-risk, high-return reward plan in which the difference in reward between correct and incorrect answers is large, and a low-risk, low-return reward plan in which there is little difference between correct and incorrect answers~\cite{sakurai13ability}.

In the weighted majority voting method based on worker ability, the parameters representing worker ability are estimated by having workers carry out work on tasks for which the correct answers are known in advance. In practice, there are many cases in which it is not possible to prepare enough tasks with correct answers, or the worker has just started working. Therefore, it is desirable to have a method that simultaneously estimates the worker's ability and the correct answer from the results of their work on tasks for which the correct answer is not known. One method uses a latent class model that alternately estimates the correct answer and the worker's ability parameters~\cite{Dawid1979}. In fact, the latent class model was proposed to derive a more reliable diagnosis from the results of medical diagnoses by multiple doctors long before the advent of crowdsourcing, and only recently has been used in crowdsourcing research to derive correct answers from multiple worker responses. The EM (Expectation Maximization) method, which is commonly used for parameter estimation in models with hidden variables (correct answers), was used in this study. Additionally, several Bayesian extensions of latent-class models have been proposed to address a small number of worker labels and complex relations among workers. For example, Community BCC~\cite{Venanzi2014-fx} considers group structures within workers, and Enhanced BCC~\cite{Li2019-EBCC} models the correlation between workers.

\subsubsection{Considering the cognitive bias of workers}
Efforts have been made to improve the quality of human computation by focusing on the characteristics of human information processing. Cognitive bias is a decision-making bias that occurs systematically because of the tendencies and limitations of human cognitive functions. Naturally, the effects of cognitive bias are also expected to appear in human-driven computations. Indeed, in crowdsourcing, which is often a platform for human computation, human cognitive bias is known to have a significant impact on the results~\cite{draws2021checklist}.

Newell and Ruths~\cite{newell2016} showed that in a repetitive image-labeling task, the crowd worker's previous task influences the results for the current task. This can be considered as a type of anchoring bias. Eickhoff~\cite{eickhoff2018} demonstrated the existence of various cognitive biases (ambiguity effect, anchoring, bandwagon effect, and decoy effect) in crowdsourcing tasks through experiments.

The individual nature of the crowd worker also influences the propensity for cognitive bias. For example, a confirmation bias is a type of cognitive bias in which workers attach importance to information that supports their hypotheses and beliefs. In crowdsourcing, workers' personal beliefs and backgrounds have also been observed to influence the outcome of their judgments~\cite{barbera2020,coscia2020}. Demartini~\cite{demartini2019} confirmed that each worker's attributes affect the outcome of a fact-checking task. Leniency and Halo errors~\cite{balzer1992} arise from the difficulty for raters to independently verify multiple rating axes. The presence of these biases in impression~\cite{biel2014} and mutual ratings~\cite{kulkarni2014} has also been noted in crowdsourced annotations.

The above studies primarily suggest the existence of negative effects of cognitive biases on task results and the need to remove these biases. Typical ways to mitigate these biases involve ingenious task design and interventions in the annotation process. Eickhoff and de Vries~\cite{eickhoff2013} explored various factors for reducing cheat submissions in crowdsourcing tasks. They reported that rewriting the task to avoid repetition or reducing the batch size alone reduced the rate of cheating. Hube et al.~\cite{hube2019} instructs participants to consider social opinions in subjective-judgement tasks and to instruct people not to bring their own potential biases into the task. By making people aware of possible biases, they are effectively mitigated. On the other hand, Duan et al.~\cite{duan2020does} also reported that workers with different perspectives did not necessarily reduce bias when they worked together.

Faltings et al.~\cite{faltings2014} showed that monetary incentives affect worker bias and proposed a bonus-adding scheme based on game theory. Gadiraju et al.~\cite{gadiraju2017self} used a pre-screening task that required workers to submit self-assessments as well as responses to a task, aiming to identify competent workers based on the Dunning-Kruger effect. Gamalmaz and Min~\cite{ijcai2021-238} proposed a method that introduced the degree to which each worker is affected by a confirmation bias in a label-integration model, which allowed for integration that eliminated the influence of bias. Echterhoff et al.~\cite{echterhoff2022ai} proposed a method to remove anchoring bias, whereby a decision on the same item will differ depending on one's past decisions, by detecting it in the data or by controlling the order in which the items are presented.

\subsubsection{Considering the relationship between tasks and workers}
In the latent-class model introduced in Section 2.1.1, worker ability was estimated by assuming that the difficulty of the problem was constant. However, to assess worker ability accurately, it is necessary to consider the difficulty of a task. The item response theory used in the test design~\cite{linden1997handbook} can be used to estimate worker ability using a task in which the correct answer is known. In crowdsourcing, Whitehill et al.~\cite{whitehill2009} proposed a method that can simultaneously determine the worker's ability and the correct answer while considering the difficulty of the task, even when the correct answer is not given. Bachrach et al.~\cite{bachrach2012grade} proposed an Bayesian extension called joint difficulty-ability-response estimation (DARE) model.

The models described thus far assumed that workers could be represented by a one-dimensional ability parameter and tasks by a one-dimensional difficulty parameter. However, human ability is multifaceted and includes various factors, such as language ability, computational ability, and memory. On the other hand, tasks also differ in terms of the abilities required to perform them, with some tasks requiring a high level of language skills but not computational skills, and other tasks requiring a high level of computational skills but not language skills. This leads to differences in worker-task ``compatibility,'' such that a task that is easy for one worker (high language ability but low computational ability) may be difficult for another worker (low language ability but high computational ability). To incorporate such situations into the model, it is necessary to represent the worker and task parameters as multi-dimensional feature vectors rather than one-dimensional parameter. Welinder et al.~\cite{welinder10muiti} proposed a method for estimating worker and task feature vectors from worker responses, using a model in which worker responses are determined probabilistically given the above worker and task feature vectors.

Another approach is to directly ask the worker about the difficulty of the task or the worker's confidence in his/her answer, rather than using computationally estimated task difficulty. This approach may be more in line with the idea of human computation in that it makes active use of human metacognitive abilities. Although there is a correlation between confidence and the rate of correct answers, there are many overconfident workers whose confidence is higher than the rate of correct answers, and many underconfident workers whose confidence is lower than the rate of correct answers. The accuracy of confidence judgments differs from person to person, and it is necessary to consider differences in the accuracy of workers' self-assessments, rather than treating the confidence level given by each worker uniformly. Therefore, by extending the latent-class model and introducing a parameter indicating the accuracy of a worker's self-reported confidence, a model that can derive the correct answer with good accuracy has also been proposed~\cite{oyama13accurate}.

To obtain quality work results, it is important to ensure that the task is performed by a worker who has the necessary competence for the task. Task recommendation formulates task assignments as a recommendation problem: Ambati et al.~\cite{Ambati2011} predicts preferred tasks based on worker attributes and other information through supervised learning; Yuen et al.~\cite{yuen2012taskrec} recommends tasks using task browsing history and past tasks; Kulkarni et al.~\cite{kulkarni2012mobileworks} also introduced a mechanism for other workers to recommend tasks. Li et al.~\cite{wom} used a model that predicted task accuracy based on worker attributes such as nationality, education level, gender, major, and personality test scores to identify appropriate workers. In addition, there have been reports of application accuracy drops in tasks dealing with language, even for the same English task, when country and speaker attributes differ between the data-collection task and the resulting application domain. Bender and Friedman~\cite{bender2018data} suggested utilizing data statements that include speaker and annotator demographics for data dealing with language.

The methods described so far assume that batch processing is performed after all workers have completed their work on all tasks, and that the degree of redundancy, i.e. the number of workers requested to perform the same task, for each task is predetermined. However, because task execution involves both human and financial costs, it is desirable to minimize redundancy if the quality can be guaranteed. For this purpose, instead of allocating the same redundancy to all tasks, it is necessary to distinguish between tasks with different levels of answer uncertainty. For tasks in which the correct answer is mostly clear from the worker's response, no further requests are made to the worker. Contrary, for tasks in which the correct answer is uncertain, the worker's response is continuously collected until the correct answer is assured. To this end, Sheng et al.~\cite{sheng2008get} proposed a method of Bayesian estimation of the probability that the answer decided by the majority vote is incorrect, which is then used to calculate the uncertainty of the answer.

Sheng et al.~\cite{sheng2008get} assumed that the abilities of all workers are equal. However, workers have different abilities, and it is possible to improve the efficiency of quality control by considering these differences when assigning tasks. As previously mentioned, worker ability can be evaluated using tasks in which the correct answers are known. However, if the number of tasks performed by a worker is small, the worker's ability remains uncertain. For example, we cannot assume that a worker's ability is low only because of his/her first few failures.

In the problem of selecting a task to collect responses, workers were preferentially assigned to tasks with high uncertainty levels. On the other hand, in the problem of selecting which worker to request, assigning a task to a worker with high uncertainty in his/her ability is not a good method. Rather, if a worker is known to have high ability, it is more efficient to continuously assign tasks to that worker. However, if we assign tasks to the same workers only, we may omit a high-capable worker who has not yet been assigned many tasks. (Conversely, there is no need to assign tasks only to those who are certain to have low ability.) This is the so-called trade-off between exploitation and exploitation and appears in various decision-making problems when dealing with uncertainty. To address such problems, a method called {\bf interval estimation}~\cite{kaelbling90learning} was introduced for crowdsourcing and used for worker selection~\cite{Donmez2009}.

\subsubsection{Quality control for other types of tasks}
There are many forms of human computation output, not just the simple labeling of data. Ranking is the ordering of given items and is used in a variety of tasks, such as evaluating search engine results. Chen et al.~\cite{chenWSDM2013} proposed a method to determine the overall ranking while considering the work quality when pairwise comparisons between two items were provided during crowdsourcing. Furthermore, they generalized traditional active learning and proposed a method for selecting pairs to be assigned to workers while considering worker quality, uncertainty in the order of pairs, and uncertainty in the model. This study determined a single ranking for all items. In contrast, Matsui et al.~\cite{matsui14crowdordering} proposed a method for simultaneously determining multiple rankings from worker ordering data for three or more items in a task, such as sorting words in a sentence.

The quality control studies described thus far have dealt with cases in which the worker produces structured outputs, such as classification and ranking. In human computation, many tasks such as writing articles also deal with unstructured outputs. For example, Wu et al.~\cite{wu2012sembler} proposed an aggregation method of sequence labels that assumed a sequential noise model. Wang and Dang~\cite{wang2022generative} proposed a sentence integration model based on a language generation model using a transformer. On the other hand, Baba and Kashima~\cite{baba13statistical} proposed a general quality control model consisting of two stages: producing unstructured outputs and evaluating them. They introduced a probabilistic generative model for quality control that considered the author's ability and evaluator bias.

\subsection{Availability}
In a traditional RAS, availability is calculated using the system uptime relative to the total time. Because human labor is the bottleneck in increasing availability in human computation, various studies on worker availability have been conducted. In the following subsections, we describe existing research on monetary and non-monetary incentives, worker behavior prediction, and crowdsourcing contests to achieve availability. Table~\ref{table:ras-availability} summarizes the related studies on the availability of human computation.

\begin{table}[t]
  \caption{Studies on availability in human computation.}
  \label{table:ras-availability}
  \centering
  \small
  \begin{tabular}{llr}
    \hline
    Category  & Reference  &  Description  \\
    \hline \hline
    Monetary incentives & Mason \& Watts (2009) \cite{mason2009financial}   & Increasing monetary incentives\\
                & Miao et al. (2022) \cite{miao2022dynamically} & Dynamic pricing of tasks\\
                & Bigham et al. (2010) \cite{bigham2010vizwiz} & Recruiting workers for real-time service\\
                & Bernstein et al. (2011) \cite{bernstein2011crowds}     & Recruiting workers for real-time service\\
                & Bernstein et al. (2012) \cite{bernstein2012analytic} & Estimating waiting time by queueing theory\\
                & Oka et al. (2014) \cite{oka2014predicting} & Giving incentive to work during declared period\\            
                & Bacon et al. (2012) \cite{bacon2012predicting} & Having workers declare their effort and completion time\\
                & d'Eon et al. (2019) \cite{deon2019paying} & Rewarding collaborative work\\
                \hline
Non-monetary incentives    & von Ahn \& Dabbish (2008) \cite{vonAhn08designing} & Introducing games with a purpose\\
    & Ipeirotis \& Gabrilovich (2014) \cite{ipeirotis2014quizz} & Recruiting volunteer workers through advertisements\\
& Dai et al. (2015) \cite{dai2015and} & Providing occasional entertainments for workers\\
\hline
Worker behavior prediction                & Cheng et al. (2019) \cite{cheng2019frog} & Worker availability prediction using machine learning\\
                & Mao et al. (2013) \cite{mao2013stop} & Predicting volunteer worker participation\\
\hline          
Crowdsourcing contests & DiPalantino \& Vojnovi\'{c} (2009) \cite{dipalantino2009crowdsourcing} & Relationship between rewards and participation in contests\\
                & Truong et al. (2022) \cite{truong2022efficient} & Selecting the optimal incentive design\\
                & Feyisetan \& Simpert (2019) \cite{feyisetan2019beyond} & Introducing contests to microtask crowdsourcing\\
                \hline
   \end{tabular}
\end{table}

\subsubsection{Monetary incentives}
One way to increase the availability of the workforce is to increase the reward, which is known to increase the quantity of work, although this does not necessarily improve the quality of the work~\cite{mason2009financial}. Setting appropriate rewards is critical to ensuring worker availability. If the reward is too low, it will be difficult to attract enough workers, whereas if the reward is too high, it will be less profitable for the client. Miao et al.~\cite{miao2022dynamically} used a deep time series model to predict the relationship between bonus and task quality and proposed a dynamic pricing mechanism for tasks by introducing bonuses.

Labor availability is particularly important for crowdsourcing real-time services. VizWiz~\cite{bigham2010vizwiz} hires workers before they are needed and keeps them on standby to solve past tasks, enabling the service to answer questions about images from blind people in real time. Similarly, the Retainer Model~\cite{bernstein2011crowds} pays workers a small fee to wait for a task, and when it becomes available, the worker can start working immediately. Furthermore, Bernstein et al.~\cite{bernstein2012analytic} used queueing theory to analyze requester waiting times and costs in the Retainer Model.

Participatory~\cite{burke06participatory} and crowd sensing~\cite{ra12medusa} asks users with mobile terminals or other devices to collect data such as images, sound, and location information from real world environments, which are often used in citizen science. The difference between conventional distributed sensing and crowdsensing is that the sensors are owned by ordinary users and require their involvement in data collection. The key challenge in crowdsourcing is obtaining participants to contribute to data collection. Therefore, research is being conducted to provide incentives for participants to work according to their declared schedules~\cite{oka2014predicting}.

For long-running tasks such as software development, it is important to know the task's completion time. The problem here is that the workers themselves, who predict the completion time, can influence this time by adjusting the amount of effort. Bacon et al.~\cite{bacon2012predicting} proposed an incentive mechanism to elicit maximum effort, while allowing workers to report their expected time of completion with honesty.

In conventional crowdsourcing, workers receive payment for work completed on an individual basis. However, some tasks may require the cooperation of multiple workers. When paying for group work, paying all members the same remuneration may cause higher performing workers to withhold their efforts. d'Eon et al.~\cite{deon2019paying} compares rewards proportional to outcomes based on equity theory and Shapley values in cooperation game theory with uniform rewards. The results suggest that pay-for-performance rewards increase the effort extended by workers to a small extent.

\subsubsection{Non-monetary incentives}
Gamification or games with a purpose~\cite{vonAhn08designing} are representative methods for encouraging participation with non-monetary incentives. Games with a purpose have known templates called output-agreement, inversion-problem, and input-agreement, which can be used to transform a task into a game with a desired outcome. Furthermore, the authors stated that timed responses, score keeping, player skill level, high-score lists, and randomness are important factors in making a game interesting.

When collecting specialized knowledge in areas such as the medical field, it is necessary to recruit workers who are knowledgeable in these areas. Ipeirotis and Gabrilovich~\cite{ipeirotis2014quizz} recruits the participation of workers who are not paid for their labor by using an advertising platform. By not paying rewards, they are able to target only those workers who are knowledgeable about the task, and attract only those who have the internal motivation to perform the task. Additionally, advertising campaigns can be optimized by providing feedback on worker conversions to the advertising platform. Furthermore, they found that providing feedback, such as the correct answers to tasks and performance of other participants led to continued participation.

Unlike machines, humans do not like performing monotonous tasks for extended periods of time. Dai et al.~\cite{dai2015and} proposed a method to keep workers engaged by occasionally providing them with entertainment, such as games and cartoons, as micro-diversions. Experimental results on several types of tasks have shown that introducing entertainment improves worker retention and response speed while maintaining quality. Furthermore, it is suggested that micro-diversion is more effective in complex cognitive tasks than in reflective tasks.

\subsubsection{Worker behavior prediction}
Research has also been conducted to predict the availability of workers using machine learning. For example, the crowdsourcing framework FROG~\cite{cheng2019frog} uses Kernel Density Estimation to predict future availability based on a worker's past work history. It also introduces smoothing based on information from social networks among workers to solve the cold-start problem for new participants with no work history.

In volunteer-based crowdsourcing, there is no financial incentive therefore, it is important to predict worker disengagement and provide appropriate intervention to workers who are likely to disengage. Mao et al.~\cite{mao2013stop} developed a model for predicting worker disengagement through supervised learning, based on task and worker characteristics, as well as worker activity data during task sessions.

\subsubsection{Crowdsourcing contests}
In crowdsourcing contests, the number of participants has a significant impact on the final output. There are usually multiple contests running simultaneously, and they compete for participation. DiPalantino and Vojnovi\'{c}~\cite{dipalantino2009crowdsourcing} model contests as all-pay auctions, where all bidders must pay their bids regardless of whether they win, and the relationship between prize money and the number of participants is analyzed. The results show that the number of participants is related to the number of prize winners. They theoretically show that the number of participants is logarithmically proportional to the amount of prize money, which is consistent with actual data when the target is limited to users who repeatedly use the crowdsourcing site.

Prize distribution in contests has been analyzed using game theory~\cite{moldovanu2001optimal,archak2009optimal} to determine whether prize money should be paid to the highest ranking participant only or to multiple participants. These theoretical studies show that the optimal prize distribution depends on the risk preferences of participants and the form of the cost function. Therefore, Truong et al.~\cite{truong2022efficient} formulated the incentive-selection problem of choosing the optimal design among multiple incentive designs as a multi-armed bandit problem and proposed an efficient algorithm to solve it. 

In conventional crowdsourcing of microtasks, workers are paid a fixed amount upon completion of the task. Contest elements have also been introduced into microtask crowdsourcing to accelerate the time required to complete tasks. Feyisetan and Simpert~\cite{feyisetan2019beyond} introduced a leaderboard based on the number and quality of tasks completed in microtask crowdsourcing to allow workers to compete. They showed that paying only the highest-ranked workers increased the speed of task completion, but decreased the quality of some types of tasks. In contrast, they also showed that increasing the number of workers rewarded can increase the number of completed tasks.

\subsection{Serviceability}
Serviceability indicates the ease of recovery when a system fails, and is evaluated by an index such as MTTR (Mean Time To Repair). To the best of our knowledge, no research has directly and quantitatively evaluated serviceability in human computation. Studies have been conducted on the construction of human computation that facilitates recovery from failures. Of particular importance is the introduction of knowledge developed in software engineering, which abstracts programming details and improves the ability to maintain complex systems. Table~\ref{table:ras-serviceability} summarizes related studies on serviceability in human computation.

\begin{table}[t]
  \caption{Studies on serviceability in human computation.}
  \label{table:ras-serviceability}
  \centering
  \small
  \begin{tabular}{llr}
    \hline
    Category  & Reference  &  Description  \\
    \hline \hline
    Workflow management &  Kittur et al. (2011) \cite{kittur2011crowdforge}  & Decomposing complex tasks into simple tasks\\
    & Dai et al. (2010) \cite{Dai2010} & Automated control of crowdsourcing workflows\\
    & Zhang et al. (2013) \cite{zhang2013automated} & Synthesizing optimal workflows\\
    & Kulkarni et al. (2012) \cite{kulkarni2012collaboratively} & Asking workers to decompose and integrate tasks\\
    & Kittur et al. (2012) \cite{kittur2012crowdweaver} & Visual management of complex tasks\\
\hline
Programming language    & Little et al.(2010) \cite{little2010turkit} & Imperative programming of fault-tolerant human computation\\
    & Minder \& Bernstein (2012) \cite{minder2012crowdlang}  & Programming using abstract patterns of human computation\\ 
                    & Park et al. (2012) \cite{park2012deco} & Writing data retrieval queries using an SQL-like language\\ 
                    & Morishima et al. (2012) \cite{morishima2012cylog}  & Declarative programming using a Prolog-like language\\ 
                    & Tranquillini et al. (2015) \cite{tranquillini2015modeling} & Visual programming using a business process modeling language\\
                    \hline
   \end{tabular}
\end{table}
\subsubsection{Workflow management}

In human computation, when complex tasks are performed, their workflows become more complex, therefore it is necessary to abstract them and make them easier to manage. CrowdForge~\cite{kittur2011crowdforge} automates the flow of decomposing complex tasks into simple tasks by abstracting them with a distributed computing framework similar to MapReduce. Conventional crowdsourcing workflows are complex and repetitive, making it difficult to maintain optimal workflow. TurKontrol~\cite{Dai2010} used a mathematical model for decision making to automate the control of crowdsourcing workflows. In the field of software engineering, there is a process called program synthesis, which generates a program from a problem specification; Zhang et al.~\cite{zhang2013automated} introduced this idea to crowdsourcing workflows. They proposed a method for synthesizing optimal workflows based on inference arising in the workflow structure and learning about worker performance.

Turkomatic~\cite{kulkarni2012collaboratively} allows workers to participate in the design and execution of workflows such as task division and integration, and visualizes the process so that requesters can supervise it. CrowdWeaver~\cite{kittur2012crowdweaver} is a system for visually managing complex workflows. It allows requesters to reuse existing tasks to create workflows, manage data flow between tasks and task progress, and modify workflows in real time.

\subsubsection{Programming languages}
There have been studies on programming language-level support for describing complex human-computation workflows. Turkit~\cite{little2010turkit} introduced a crash-and-return programming model that allows imperative programming to describe fault-tolerant human computation. CrowdLang~\cite{minder2012crowdlang} is a programming language for designing and implementing human computation, which can incorporate patterns that appear in human computation through abstraction, such as group decision making. One type of abstract programming language is a declarative language termed SQL, which is a standard declarative data-retrieval language. That is, the user only describes what he/she wants to retrieve from the database, and the steps for retrieval are left to the optimization process and are hidden from the user. Deco~\cite{park2012deco} is a system that allows data retrieval using crowdsourcing to be written in SQL. CyLog, a logic programming language similar to Prolog, is a declarative language for describing crowdsourcing. BPMN4Crowd~\cite{tranquillini2015modeling} is a business process model and notation (BPMN)-based modeling language. It allows crowdsourcing workflows to be intuitively described and integrated into business processes by means of diagrams using a visual editor.

%% file: sec/socialtrust.tex
In this chapter, we discuss the opposing perspective that expressed in the previous chapter, that is, the trustworthiness that a human computation system instills in its users, society, and participants (crowd workers in many cases) in human computation. Similar to the previous discussion on the trustworthiness of AI~\cite{kaur2022trustworthy}, human computation systems must be operated ethically to earn the trust of society. The processes and deliverables of human computation systems should also be ethical and explicable, such that general public can understand and accept them.

In ethical human computation systems, crowd workers must be treated fairly and respectfully. Additionally, they must be protected from risk. Examples of such requirements are protecting workers' privacy and ensuring crowd workers are treated appropriately as laborers.

Human computation systems must handle data generated by workers in a scientifically and socially appropriate manner. When a system learns from the crowd-generated data and makes decisions based on them, its behavior is expected to follow the AI ethics guidelines.  In terms of security, an AI system should ensure that socially undesirable,  malicious, or privacy-invasive tasks are not performed. Moreover, it is necessary to consider ethical requirements such as fairness, accountability, and transparency of the outcomes of the system. For human computation systems, data biases, such as sampling and annotation biases of annotators, have a significant impact on the fairness of their outcomes. The explanation of human computational processes and their outcomes is closely related to accountability and transparency. This explanation should be reasonable and understandable for non-experts. Along with the development of explanation methods, it is necessary to evaluate the understandability of explanations.

\subsection{Ethical Treatment of Crowd Workers}
The fair treatment of crowd workers is a major ethical requirements in human computation. Crowd workers often conduct subjective tasks such as data annotation and questionnaire responses. Therefore, it is necessary to protect workers from possible risk during task execution. Privacy preservation of workers, and labor management are also required.

In human computation, tasks are often based on subjective ratings or sensory evaluations conducted by humans. These should meet the same standards as the ethics of research and experiments in the behavioral sciences and psychology, which involve questionnaires and other surveys of human subjects. The Belmont report~\cite{united1978belmont} summarizes ethical principles and guidelines for research involving human subjects. The report identified three core principles: respect for persons, beneficence, and justice. In applying these principles, we need to consider informed consent, the assessment of risks and benefits, and selection of subjects. All relevant information should be provided in an understandable and accessible manner. In particular, possible risks in task execution, such as psychological invasion, must be noted. It is also required that workers should not be disadvantaged in the task assignments. All relevant studies should be reviewed and approved by the Institutional Review Board (IRB).

As human computation is invariably conducted in an online environment, we should also refer to the guidelines for participant protection during online research. Several guidelines has been proposed for participant protection in online research~\cite{frankel1999ethical,bruckman2002ethical}. To protect privacy in online interactions, the requester must protect workers' private information such as email addresses. When obtaining consent, they must be careful to provide information in a manner that the participants can comprehend. Although the online work environment has changed significantly since these guidelines were proposed, and crowd workers have become more proficient at online work, the aforementioned points should still be considered.

If a task contains deception, intentional misleading of subjects, workers must be informed. User studies related to privacy and security such as phishing unavoidably contain deception~\cite{jagatic2007social}. While research involving deception can be of great benefit to academia and product development, additional care must be taken with crowd workers. The requesters must design research protocols that minimize risk. They must also conduct debriefing sessions for workers, and the deception in the task and its necessity must be explained in an understandable manner~\cite{finn2007designing}.

The crowdsourcing work environment must be considered when implementing human computation on a crowdsourcing platform. It has been noted that, while the crowdsourcing marketplace allows workers to work without geographic or time constraints, labor exploitation can occur due to differences in position and economic disparity~\cite{silberman2010ethics}. One of the concerns raised on low compensation is that crowdsourcing of translation tasks may upset the balance of the market by competing with professional translators~\cite{dolmaya2011ethics}. Research communities currently suggest that rewards should at least be the minimum wage, and research papers should clearly state how workers are compensated. For example, Prolific, a research-purpose crowdsourcing platform, has a minimum hourly wage that encourages appropriate payments. On the other hand, a survey of workers revealed that, they are often not trained in labor management and sometimes work on several jobs at simultaneously~\cite{kaplan2018striving}, while labor management on crowdsourcing still has many problems. Human computation systems should ensure fair evaluation of workers and requesters. Mutual evaluation between clients and workers is a prominent feature of crowdsourcing. For workers to receive appropriate compensation and work opportunities, appropriate evaluation of deliverables must be considered. In addition, proper evaluation of the requesters is also important, as they may be unfairly undervalued, resulting in loss of task assignment opportunities.

These systems must avoid inappropriate tasks to ensure security of human computation. Examples of inappropriate tasks include the unauthorized acquisition of personal information and cooperation in stealth marketing and slander. System administrators must constantly monitor and remove inappropriate tasks. From the perspective of ensuring the security of human computation, the systems have to avoid inappropriate tasks. An automated monitoring system using machine learning was proposed to improve the efficiency of such monitoring~\cite{baba2013leveraging}.

The privacy of workers and tasks must be properly protected. Crowd workers' private information such as names and e-mail addresses must be secured. In addition, there is a risk of sensitive information being inferred from task execution. For example, in a mobile crowdsourcing which uses mobile devices and sensing to collect data, the workers' location can be inferred from task outcomes and the information exchanged during task execution. Feng et al.~\cite{feng2019Survey} identified privacy and security threats in mobile crowdsourcing and reviewed existing countermeasures.

Worker privacy protection by using privacy-preserving data mining techniques has also been studied. Kajino et al.~\cite{kajino2014preserving} proposed a secure computation protocol for crowd label aggregation, in which labels provided by crowd workers and their ability parameters are kept secret, but the client can obtain aggregated labels. The protocol uses homomorphic encryption schemes to perform the secure computation to estimate the true aggregated label from encrypted labels. Similarly, secret computation techniques were applied for task-worker matching~\cite{shu2018anonymous} and for task recommendation~\cite{shu2018privacy}, while protecting worker privacy.

Task instances may also contain private information. For example, in medical tasks, personal information can be inferred from images and texts of medical records. Task division is an approach for coping with such risks. For text privacy, Little et al.~\cite{Little2011} proposed a method that divides medical records and presents them to different workers during transcription tasks. To protect image privacy,  Kajino et al.~\cite{kajino2014instance} investigated the properties of the instance clipping protocol that clips the images by a window of a fixed size. 

\subsection{Fairness and biases in human computation}
Accounting for the fairness of AI systems has recently become increasingly important. AI systems sometimes exhibit discriminatory behavior toward certain groups or populations in high-stakes decisions, such as hiring and personnel evaluations. Examples of cases that drew criticisms include a hiring AI systems discriminating against women~\cite{dastin2022amazon} and a facial recognition system that reported a difference in accuracy between groups of different gender and race~\cite{buolamwini2018gender}. Furthermore, a chat-bot reflected hate speeches contained in its training data~\cite{neff2016talking}.

As most AI systems and algorithms are data-driven, they reflect biases in the training data~\cite{barocas-hardt-narayanan}. For example, if the data has a sampling bias, the characteristics of a minority in the data are less reflected in the result. The labels or annotations may have biases that discriminate against certain groups or populations. Models learned from biased labels prone to reflect biases in their predictions. For natural language processing model, Shah et al.~\cite{shah2020predictive} suggested general mathematical definitions of predictive biases. They differentiated four potential origins of biases: label bias, selection bias, model over--amplification, and semantic bias. Here we focus on workers' annotation or label bias in the human computation system. Table~\ref{table:bias} summarizes the studies related to the annotation biases.

\begin{table}[t]
  \caption{Studies on annotation bias in human computation.}
  \label{table:bias}
  \centering
  \small
  \begin{tabular}{llr}
    \hline
    Category  & Reference  &  Description  \\
    \hline
    
    \hline
    Systematic annotation biases & Sen et al.~\cite{sen2015standards} & Difference in semantic relatedness judgments between communities  \\
    & Dong \& Fu~\cite{dong2012war} &  Difference in image tagging tasks between countries \\
    & Otterbacher et al.~\cite{Otterbacher2018genderbias} & Gender bias of sexists in search result evaluation \\
    \hline
    Modeling worker biases & Liu et al.~\cite{liu-etal-2022-toward} & Modeling annotator group bias in label aggregation \\
    & Davani et al.~\cite{davani-etal-2022-dealing} & Modeling each annotator using multi-task learning  \\
    \hline
    Countermeasures for biases & Barbosa et al.~\cite{barbosa2019rehumanized} & Task allocation considering the human factors of workers \\
    & Ueda et al.~\cite{Ueda22} & Enhancement of the minority labels in label aggregation \\
     \hline
    Multi-modality in annotations & Gordon et al.~\cite{gordon2021disagreement} & Classification  using disagreement-adjusted metric \\
    & Gordon et al.~\cite{gordon2022jury} &  Label aggregation for each subgroup \\
    \hline
  \end{tabular}
\end{table}

Crowd workers, who are often participants in decision-making as well as providers of data and inputs in human computation, tend to be biased in subjective tasks, depending their personal preferences or inclinations. Several studies have shown systematic annotation biases. Sen et al.~\cite{sen2015standards} showed that different communities create different gold standards in semantic relatedness judgments. Dong and Fu~\cite{dong2012war} found that European-Americans and Chinese crowd workers may provide different tags in image tagging tasks. Otterbacher et al.~\cite{Otterbacher2018genderbias} also reported that sexist workers were less likely to report gender bias in evaluating image search results.

Methods for modeling and estimating worker bias have also been proposed. Liu et al.~\cite{liu-etal-2022-toward} developed a framework to capture annotator group bias. The framework extends the probabilistic graphical model for label aggregation. It uses the probabilistic graphical model with additional variables representing annotator group bias. Davani et al.~\cite{davani-etal-2022-dealing} investigated a multi-task learning approach that treats the prediction of each annotator' judgments as separate subtasks, while sharing a common learned representation of the subtasks.

The presence of biases has a significant impact on the outputs of the AI systems trained on them, which can be a major social risk, for example, discrimination against certain groups or populations. In order to deal with such biases, there is an attempt to assign workers with an appropriate demographic composition for given tasks. Barbosa et al.~\cite{barbosa2019rehumanized} proposed a framework that allocates tasks considering the human factors of workers. The frameworks translate the task assignment process into a multi-objective optimization problem. By routing tasks to workers based on demographics, it mitigates biases in the worker sampling. On the other hand, Ueda et al.~\cite{Ueda22} viewed this as a correction problem of sampling biases in worker recruitment, and enhanced the minority in the data in the results of label aggregation. 

Certain subgroups of crowd workers may be biased in subjective tasks such as detecting emotion, aggression, and hate speech, which often reflects their preferences or inclinations. Such biases may result in annotation disagreements in label aggregation. However, such disagreements do not necessarily indicate low-quality annotation. Alm~\cite{ovesdotter-alm-2011-subjective} identified potential challenges for subjective tasks including the absence of the grand truth. Alm also discussed the implications. Evaluation techniques deserve careful thought if the ground truth needs to be reassessed. Inter-annotator agreement schemes may not be appropriate and divergence and variation in annotation may provide a useful understanding of the task.

The presence of subgroups with multiple systematic differences can lead to multi-modality in their opinions. Gordon et al.~\cite{gordon2021disagreement} proposed an approach that separates bias from noise and uncertainty, rather than aggregating annotation labels into a single ground truth label in situations where there is multi-modality in the opinions. Although there is currently no standardized method for determining which subgroups' opinions are more important than the others, Gordon et al.~\cite{gordon2022jury} also proposes ``jury learning'' that presents aggregated labels for each subgroup defined by their attributes.

Note that fairness is sometimes a trade-off with privacy. It is often necessary to use the worker's attributes to deal with biases in human computation systems. However, workers' demographic or profiled information is sometimes privacy-sensitive. For example, the GDPR states that when profiling based on personal data, it is necessary to inform the data provider of the purpose and use of the data~\cite{voigt2017eu}. Appropriate notices must be provided to crowd workers. Also, privacy-sensitive information must be protected in the human computation systems.

\subsection{Transparency of human computation systems}
Along with the recent spread of AI, there has been a growing demand for explaining AI. To gain the trust of the users, it is necessary to explain the behavior of AI systems~\cite{ribeiro2016should}. However, what should be explained depends on each individual case~\cite{adadi2018peeking}.

To ensure the transparency of human computation systems, in addition to the explanation of AI, profiles and characteristics of data produced by workers is important. Besides, explanation of AI should be understandabe and justified for general users. This section covers studies on explanation of the data and evaluation of interoperability by humans. The related studies are summarized in Table~\ref{table:transparency}.

There have been several proposals for explaining data. Bender and Friedman~\cite{bender2018data} proposed data statements for natural language processing. Data statements were proposed to address scientific and ethical issues that result from the use of data from certain populations for other populations. A data statement is a characterization of a dataset that provides a context of the dataset for developers and users. Gebru et al.~\cite{gebru2021datasheets} proposed datasheets for datasets. A datasheet describes characteristics of datasets for machine learning. It documents its motivation, composition, collection process, recommended uses, and other information. Gebru et al. provided a set of questions designed to elicit the information and a workflow for dataset creators to use when answering these questions. Note that privacy risks in these explanation should be considered. There is a risk of privacy leakage in the explanation of data creation, for example, by disclosing detailed profiles of a small number of crowd workers~\cite{bender2018data}.

\begin{table}[t]
  \caption{Studies on transparency in human computation.}
  \label{table:transparency}
  \centering
  \small
  \begin{tabular}{llr}
    \hline
    Category  & Reference  &  Description  \\
    \hline\hline
     Data explanation & Bender \& Friedman  \cite{bender2018data} & Dataset description for natural language processing  \\
    & Gebru et al. \cite{gebru2021datasheets} &  Dataset description for machine learning \\
    \hline
    Evaluation of interpretability & Doshi-Velez \& Kim~\cite{doshi2017towards}& Classification of crowdsourced evaluation schemes \\
    & Hutton et al.~\cite{hutton2012crowdsourcing} & Worker preference of explanations for text classifiers  \\
    & Selvaraju et al.~\cite{selvaraju2017grad} & Worker rating of relative reliability of the visual explanations \\
    & Jeyakumar et al.~\cite{jeyakumar2020can} & Comparison of user preferences for the explanation methods \\
    & Can et al.~\cite{can2018ambiance} &  Evaluation of visual ambiance cues \\
    & Lu et al.~\cite{lu2021crowdsourcing} & Evaluation based on a human computation game \\
    \hline
  \end{tabular}
\end{table}

To explain machine learning models and their outcomes, various interpretation methods have been proposed~\cite{guidotti2018survey}. Major approaches of interpretation methods include local explanation, global explanation, and example-based explanation. LIME~\cite{ribeiro2016should} is a representative local explanation method. LIME estimates local linear models for explanation of a certain outcome. In the case of an image classifier, one approach of local explanation is to identify pixels that strongly influence the outcome~\cite{baehrens2009explain}.

Although the aforementioned explanation methods provide explanations and interpretations of algorithmic decision processes, it is not clear whether a certain explanation is appropriate, suitable, and reasonable for a certain task. Several methods have been proposed to automatically evaluate the performance of explanation methods. For example, Samek et al. proposed a methodology based on region perturbation for evaluating ordered collections of pixels such as heatmaps~\cite{samek2016evaluating}. However, such computational evaluation of explanation is not necessarily the same for humans' evaluation~\cite{narayanan2018humans}. Rationalization provided by machine learning algorithms can be difficult to interpret. For example, algorithms may recognize objects based on their relationship to the background rather than the object itself. An airplane may be recognized by the sky, because the background of the image of an airplane is often the sky. This is a reasonable strategy for algorithms to make decisions based on statistical information. However, this is not justified for humans. Verifying the rationality of the explanation can be effective to build interpretable model for humans.

It is reasonable to use crowdsourcing as a basis for scaling up human-based evaluation of interpretability. Several human-based evaluation schemes using crowdsourcing have been proposed to measure the capability of explanation methods. In the classification by Doshi-Velez and Kim~\cite{doshi2017towards}, crowdsourced evaluation schemes can be divided into three types: binary forced choice, forward simulation/prediction, and counterfactual simulation. Hutton et al.~\cite{hutton2012crowdsourcing} assessed the explanations for text classification by using crowd worker evaluation. The workers compared human-- and computer--generated explanations and indicated which they preferred and why. They demonstrated a slight preference for computer-generated explanations. Selvaraju et al.~\cite{selvaraju2017grad} conducted user studies to measure the reliability of the visual explanation for image classification. Workers were instructed to rate the reliability of the models relative to each other. Jeyakumar et al.~\cite{jeyakumar2020can} performed a cross-analysis user study to compare the explanation methods to assess user preferences. The study was conducted across applications spanning image, text, audio, and sensory domains. Can et al.~\cite{can2018ambiance} conducted crowd-based assessment of machine recognition of ambiance. Lu et al.~\cite{lu2021crowdsourcing} proposed a human-based evaluation method based on Peek-a-boom~\cite{von2006peekaboom}, a human computation game used for image annotation. In Peek-a-boom, the Boom player selects important parts of an image and presents them to the Peek player. An XAI method plays the Boom instead of another human.

%% file: sec/human-ai.tex
\begin{figure}[t]
  \centering
  \includegraphics[width=1\linewidth]{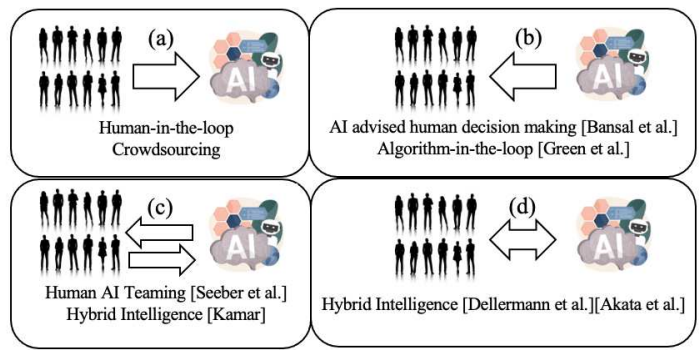}
  \caption{Collaborative human computation can take several forms: (a) humans intervene and participate in decision-making of AI. This includes human-in-the-loop. (b) AI supports human decision-making. This includes AI advised human decision-making and algorithm-in-the-loop. (c) humans and AI support each other and sometimes work together as a team to solve problems. This includes human--AI teaming. (d) humans and AI co-evolve and build mutually reciprocal relationships. This includes Hybrid Intelligence.}
  \label{fig:collaboration}
\end{figure}

Chapters 2 and 3 discussed the one-way trust between humans and AI.
In this chapter, we consider two-way trust, or human--AI collaboration, in which humans and AI build reciprocal relationships and work together to accomplish difficult tasks (which cannot be done by one or the other in isolation).
Bedwell et al.~\cite{bedwell2012} organized the cross-disciplinary concept of collaboration, with ``evolving process whereby two or more social entities actively and reciprocally engage in joint activities aimed at achieving at least one shared goal'' as the definition of collaboration. They define collaboration as a superordinate construct, which is distinguished from related constructs, such as teamwork, coordination, and cooperation, under their evaluation criteria and definition components of ``evolving process,'' ``two or more social entities,'' ``actively and reciprocally participate,'' and ``achieving at least one shared goal.''
Their definition of collaboration depicts the ideal form of collaborative human computation for solving complex problems.
In the process of reaching such an ideal goal of collaborative human computation, various forms of human--AI collaboration in the context of human computation have been studied (Fig.~\ref{fig:collaboration}). In the following sections, we will overview each form of human--AI collaboration by empathizing its associated trust aspects.

\subsection{Human-in-the-loop AI/Machine Learning}
Human-in-the-loop AI is a form of human--AI collaboration in which trustworthiness is ensured by human participation in part of the workflow (in which AI takes the initiative). In this human-in-the-loop process, humans play a partial role by providing information for AI to make decisions or by intervening and participating in decision-making by confirming or modifying the decisions of AI.

An early attempt to involve humans in the AI/machine learning process was VizWiz~\cite{bigham2010vizwiz}, a human-in-the-loop visual question answering application to assist visually impaired persons.
VizWiz uses the recognition results of the AI as-is when the AI is sufficiently confident, but automatically issues crowdsourcing tasks that leave the decision to the human when it is not.
In addition, Wilder et al.~\cite{wilder2021learning} proposed a method to quantify the cost of querying a human and the cost of making an incorrect decision using AI alone, and to make an overall decision that balances these costs.
A more complex case is a workflow such as content generation, which consists of various steps such as generation, modification, and evaluation.
AI controls the workflow, which consists of multiple humans performing various tasks, so that it is executed in an overall optimal manner.
Dai et al.~\cite{Dai2010,Dai2011} considered workflow control as a Markov decision process and proposed an approach for optimally controlling this process.

In human-in-the-loop machine learning, humans participate in certain parts of the machine learning process, such as providing data and data modeling.
Providing annotations for target labels in supervised learning is a typical example of this.
For example, Raykar et al.~\cite{Raykar2010} proposed the learning from crowds setting that simultaneously estimates ground truth target labels, worker confidence, and prediction models based on annotations provided by crowd workers with different (unknown) reliability levels. from crowds setting.
We will not cover all of them here, but there are developments such as active learning~\cite{Yan2011}, which actively selects unlabeled data and workers to request annotations to reduce the total annotation costs, and extensions to deep learning~\cite{rodrigues2018deep}.

What can be expected from human contributions is not limited to providing target labels, but also includes these input features for machine-learning prediction models.
For example, Branson et al.~\cite{branson2010visual} proposed a human-in-the-loop image recognition system in which humans were responsible for extracting (abstract) features for the inputs of prediction models.
This approach is effective in difficult tasks where sufficient data are not available for AI to be able to identify images, while even typical humans cannot directly recognize target labels owing to the lack of expertise.
Cheng and Bernstein~\cite{cheng2015flock} proposed the idea of asking humans to define features through tasks that ask them to distinguish between positive and negative examples.
Furthermore, Takahama et al.~\cite{takahama2018adaflock} proposed a boosting-based method that adaptively and efficiently performs the training.

Another approach is to obtain a representation of the data from the similarity evaluations of the data provided by humans.
Gomez et al.~\cite{Gomes2011} proposed a Bayesian method for obtaining data embeddings that reflect the results of similarity comparisons of pairs of data by crowd workers.
For humans, relative evaluation is often easier than absolute evaluation.
That is, it is easier to answer the relative similarity question ``Is object A more similar to object B or object C?'' than the absolute similarity question ``Are objects A and B similar?''
Tamuz et al.~\cite{Tamuz2011} and Wilbert et al.~\cite{wilber2014cost} obtained object similarity and embedding from these triplet-wise comparisons.
Furthermore, Amid and Ukkonen~\cite{amid2015multiview} proposed a multi-view embedding that considered comparisons from several different viewpoints.

The execution of data analysis processes using machine learning algorithms by a large number of humans can also be considered a human-in-the-loop process in a broad sense.
A typical example is data analysis competitions such as Kaggle.
In actual data modeling, no single method always achieves the best performance~\cite{wolpert2002supervised}, and it is very effective to search extensively for a model that fits data using many people.
Typical competitions employ winner-take-all competitive mechanisms, that does not motivate cooperation among participants.
To overcome this problem, a competition mechanism has been proposed that allocates rewards linked to the performance gains brought by participants in order to motivate them to contribute and share their models early~\cite{abernethy2011collaborative}.

Kamar et al.~\cite{kamar2016directions} defined hybrid intelligence as a human--AI collaboration that integrates human knowledge into AI systems to complement AI capabilities, such as human-in-the-loop.
In the hybrid intelligence, AI systems can delegate tasks to humans if necessary.
Human intervention can prevent the errors that AI systems alone may cause. 
Human feedback also improves this cycle, allowing AI systems to learn continuously. 
Therefore, AI systems have to decide when and how to benefit from complementation of human knowledge. 
Such decisions require AI systems to be equipped with models based on their and humans's capabilities, availabilities, and costs. 
As components of the hybrid intelligence, humans can serve a wide range of roles, from workers in crowdsourcing and citizen science to teachers who intervene more actively in AI systems.
In active intervention in AI systems, hybrid human--AI teamwork, such as human--AI teaming discussed below, becomes essential, where AI is required to determine how to support humans in successful collaboration as a teammate.

In contrast to human-in-the-loop, where humans participate in AI and machine learning processes, 
the idea of algorithm-in-the-loop, where AI and machine learning models support human decision-making, has also been proposed.
Green et al.~\cite{green2019principles} propose three behavioural principles of algorithm-in-the-loop: ``people using the algorithm should make more accurate predictions than they could without the algorithm'' (accuracy), ``people should accurately evaluate their own and the algorithm's performance and should calibrate their use of the algorithm to account for its accuracy and errors'' (reliability), ``people should interact with the algorithm in ways that are unbiased with regard to race, gender, and other sensitive attributes'' (fairness).

In relation to the algorithm-in-the-loop to support human decision-making, the idea of AI-advised human decision-making, where humans and AI make decisions as a team and solve complex problems, has also been proposed as a form of human--AI collaboration~\cite{bansal2019updates}.
In AI-advised human decision-making, a user receives action recommendations (predicted results in a simple case) from the AI system to solve problems.
The user then executes the proposed action of the AI or takes a different action in some cases.
Such decision-making can be applied to decision-critical areas, such as medical diagnosis, recruitment selection, and loan approval.

\subsection{Human--AI Teaming}
Regarding human--AI teaming as a form of collaboration, where humans and AI work together as a team to solve problems, Seeber et al.~\cite{seeber2020} summarise their research agenda based on research questions obtained from a survey of researchers in the field of collaboration studies.
They considered a hypothetical scenario, in which a devastating hurricane hits a small country.
In such complex situations, machines, as teammates, are assumed to collaborate with humans to solve complex and diverse problems.
The scenario embodies a complex situation in which human and AI teammates quickly analyse the situation, communicate and cooperate, coordinate emergency responses, and find reasonable solutions.
Collaborative problem-solving between humans and AI requires identifying the causes of the problems, proposing and evaluating solutions, selecting options, planning, acting, learning from past interactions, and participating in after-action reviews.
Therein, they define machines as teammates as ``those technologies that draw inferences from information, derive new insights from information, find and provide relevant information to test assumptions, debate the validity of propositions offering evidence and arguments, propose solutions to unstructured problems, and participate in cognitive decision-making processes with
human actors''.
They designed machines as teammates from seven perspectives: appearance, sensing and awareness, learning and knowledge processing, conversation, architecture, visibility, and reliability.

Zhang et al.~\cite{zhang2021ideal} studied the relationship between humans and AI as teammates in a multiplayer online game.
In particular, they addressed three research questions: (1) how people perceive AI teammates; (2) what factors influence people to team up with AI; and (3) what people expect from their AI teammates. Their results demonstrated that the perceptions of AI teammates included both positive and negative perceptions. AI is often perceived as a tool, past experiences and attitudes in collaboration with AI affect people's willingness to team up with AI, and AI teammates are expected to have proficient skills, share common understandings with people, communicate with people.

\subsection{Hybrid Intelligence}
As extensive discussions of human--AI collaboration, where humans and AI work together as a team to solve problems, several studies propose a new form of human--AI collaboration in which both humans and AI co-evolve and build mutually reciprocal relationships.
Dellermann et al.~\cite{dellermann2019future} describe such co-evolving and reciprocal relationships between humans and AI as ``Systems that have the ability to accomplish complex goals by combining human and artificial intelligence to collectively achieve superior results than each of the entities could have done in separation and continuously improve by learning from each other''. They refer to such systems as the hybrid intelligence (note that its definition differs from that in Kamar et al.~\cite{kamar2016directions}). They conducted an extensive literature review based on search queries related to the hybrid intelligence system\footnote{``hybrid intelligence OR human-in-the-loop OR interactive machine learning OR machine teaching OR machine learning AND crowdsourcing OR human supervision OR human understandable machine learning OR human concept learning''}. Consequently, they organized the knowledge for designing the hybrid intelligence system based on four meta-dimensions, 16 sub-dimensions, and 50 categories (Table~\ref{tab:taxnobyofHI}).

\begin{table}
  \caption{Taxonomy of hybrid intelligence design~\cite{dellermann2019future}}
  \label{tab:taxnobyofHI}
  \begin{tabular}{l|l}
    \hline
    Meta-dimensions & Dimensions\\
    \hline
    \hline
    Task Characteristics & type, goals, data representation, timing\\
    Learning Paradigm   & augmentation, machine learning, human learning\\
    Human--AI Interaction & machine teaching, teaching interaction, expertise requirements,\\
     & amount of human input, aggregation, incentives\\
    AI-Human Interaction & query strategy, machine feedback, interpretability\\
   \hline
\end{tabular}
\end{table}

Akata et al.~\cite{akata2020} define the Hybrid Intelligence, with particular focus on AI amplifying human intellectual capabilities, as ``the combination of human and machine intelligence, augmenting human intellect and capabilities instead of replacing them, to make meaningful decisions, perform appropriate actions, and achieve goals that were unreachable by either humans or machines.'' They organized research questions for the hybrid intelligence system in terms of collaborative, adaptive, responsible, and explainable aspects.

AlphaGO exemplifies the potential of the hybrid intelligence systems. In AlphoGO, AI learns from a vast amount of game data, and it also amplifies human knowledge by bringing the novel knowledge of Go to human players.
In this way, humans and AI in the hybrid intelligence learn from each other through various mechanisms, such as labelling, demonstrating, teaching adversarial moves, criticizing, and rewarding. 
They can then co-evolve and establish reciprocal relationships. The complex problems solved by the hybrid intelligence are characterized as time-variant, dynamic, requiring considerable domain knowledge, and having no specific ground truth~\cite{dellermann2019hybrid}.
Application areas of the hybrid intelligence include strategic decision-making, 
science, healthcare, education, innovation and creativity~\cite{dellermann2019hybrid,akata2020}.

\subsection{Trust in Collaborative Human Computation}
In human--AI collaboration, such as the hybrid intelligence described earlier, establishing appropriate trust between humans and AI is crucial.
In particular, Dellermann et al.~\cite{dellermann2019hybrid} discussed the interpretability and transparency of AI as the foundations of trust.
They argued that interpretability is important for removing bias, achieving reliability and robustness, causality of learning, debugging learning, and establishing trust. In particular, they argued that interpretability can be achieved through (1) transparency that allows the black box of algorithms to open; (2) global interpretability, which provides the general interpretability of machine learning models; and (3) local prediction interpretability, which makes more complex models interpretable for a single prediction.
In a related discussion, Vossing et al.~\cite{vossing2022} discussed that explanations (descriptions of reasoning processes)~\cite{gilpin2018explaining} that represent the internal states of the system in an understandable manner could help build trust. 
In human--AI collaboration, we cannot fully understand AI decisions. 
Therefore, they also discussed the importance of establishing trust for dealing with the complexity and uncertainty of AI systems.

When humans and AI work together as a team to solve problems, building trust significantly impacts reasonable decisions on when to follow recommendations from AI.
Regarding when to trust AI and its recommendations, Bansal et al.~\cite{bansal2019updates} introduced mental models of AI.
Concerning this discussion on mental models and trust (particularly learned trust~\cite{marsh2003role} based on contexts and past experiences), a previous study in cognitive psychology shows that people build mental models when interacting with complex systems, which facilitate their use of the system~\cite{norman1988psychology}.
Another study showed that people tend to build mental models even for autonomous systems such as AI agents~\cite{kulesza2012tell}.
Furthermore, Hoff and Bashir~\cite{hoff2015trust} discussed the relationship between mental models and trust in systems.
As an extension of these discussions, Bansal et al.~\cite{bansal2019updates} argued that building sound mental models of AI helps humans decide when to trust AI, which in turn leads to improving human--AI team performance.

Bansal et al.~\cite{bansal2019beyond} also argued that maximizing AI accuracy does not necessarily lead to maximising human--AI team performance in human--AI collaboration.
They discuss the importance of accessing when and how humans and AI should complement each other to improve human--AI team performance.
In relation to the earlier discussion of mental models, they also proposed building mental models of error boundaries regarding when AI makes errors, which allows people to decide when to follow recommendations from AI appropriately.
The importance of mental models has also recently been recognised in human-in-the-loop systems~\cite{chakraborti2018algorithms} and human-computer interactions~\cite{kaur2019building}.

\subsection{Explainable AI in Collaborative Human Computation}
A previous study has showed that building more accurate mental models of the system, in general, enhances trust in the system through explanations~\cite{staab2002intelligent}.
Regarding explanations and trust in human--AI collaboration, Lai et al.~\cite{lai2019human} investigated whether AI explanations improved human decision-making performance in tasks of deception-detection.
Their experimental results demonstrated that human performance improved slightly when AI presented only explanations, without predicted labels.
They argued that there is a trade-off between human performance and independence in AI-assisted decision-making. Explanations from AI may mitigate this trade-off.
In line with this discussion, Lai et al.~\cite{lai2021towards} further conducted a comprehensive survey on human--AI decision-making.
In the survey, they categorize the elements of explanations from AI (called ``AI assistance elements'') as ``Prediction'', ``Information about predictions'', ``Information about models (and training data)'', and ``Other'' (Table~\ref{tab:AIassitelem}), which gives an overview of which explanations are used in existing studies on human--AI decision-making.

\begin{table}
  \caption{AI assistance elements in human--AI decision-making~\cite{lai2021towards}}
  \label{tab:AIassitelem}
  \begin{tabular}{l|l}
   \hline
   AI assistance elements & Examples \\
   \hline
   \hline
    Information about predictions & model uncertainty, local feature importance, rule-based explanations, \\
    & example-based methods, counterfactual explanations, \\
    & natural language explanations, partial decision boundary\\
    \hline
    Information about models (and training data) & model performance, global feature importance, \\
    & presentation of simple models, global example-based explanations,  \\
    & model documentation, information about training data\\
    \hline
    Others & level of user agency, \\
           & interventions or workflows affecting cognitive process\\
    \hline
\end{tabular}
\end{table}

Previous studies have demonstrated the effects of explanations from AI in human--AI collaboration.
Nouran et al.~\cite{nourani2019effects} studied how explanations affect people's mental models of AI.
Specifically, they investigated how people perceive the accuracy of AI in an image classification task, depending on meaningful and non-meaningful explanations by AI (highlighting the areas of an image on which predictions are based).
Their experimental results showed that people estimated the accuracy of AI with non-meaningful explanations lower.
They argued that if explanations are not based on rationality that people can understand, they cannot accurately estimate the accuracy of AI, which leads to less trust in AI.
Smith et al.~\cite{smith2020} discussed that explanations from AI and human feedback are complementary.
They showed that the lack of feedback opportunities for explanations from AI (highlighting words on which predictions are based in their text categorization task) has a negative impact on user experiences.
They pointed out that human feedback should not be requested without explanations from the AI. In another study conducted on an image face detection task, Honeycutt et al.~\cite{honeycutt2020soliciting} reported that requesting feedback makes people perceive the system to be more inaccurate, resulting in reduced trust.
Wang et al.~\cite{wang2021} identify three desirable properties of AI explanations in the context of human--AI decision-making: (1) they improve people's understanding; (2) they help people to recognize uncertainty in the AI model; and (3) they help increase trust in the AI model.
They also pointed out that people's domain knowledge of tasks is required when receiving AI' explanations. They compare several explanatory methods, including feature importance-based, feature contribution-based, nearest neighbors and counter factual-based, in two types of human--AI decision-making tasks, recidivism prediction and forest cover prediction. Their experimental results revealed that none of the explanatory methods satisfied the aforementioned properties when people had limited domain knowledge of the tasks.

\subsection{Performance in Collaborative Human Computation}
Several side effects of explanations from AI on the performance of human--AI collaboration have been pointed out.
The first problem is that explanations from AI may lead to an information overload.
Alufaisan et al.~\cite{alufaisan2021does} showed in two types of decision-making tasks, recidivism prediction and income status prediction, that while AI's predictions improve the accuracy of human decision-making, AI's explanations (based on LIME) about the predictions have no effect on improving the accuracy.
This may be because AI's explanations lead to an information overload and therefore affect human cognitive abilities to detect errors in AI's predictions.
The second problem is that explanations from AI may cause humans to unconditionally trust AI and accept its suggestions.
Bansal et al.~\cite{bansal2021} conducted an experiment investigating the impact of explanations on human--AI team performance in two types of decision-making tasks: text classification (sentiment analysis) and question answering. 
Therein, their system highlights words as local explanations on which AI' predictions are based.
Their experimental results showed that while their subjects accepted AI suggestions thanks to their explanations, the explanations did not necessarily contribute to complementary performance improvements in human--AI collaboration. They argued that as explanations increase trust in AI, humans are accepting the AI suggestions regardless of whether they are correct or incorrect.

Regarding the impact of AI' explanations on human--AI team performance, a previous study pointed out that some issues exist when evaluating performance.
Bunica et al.~\cite{bucinca2020} argued that the inferiority of human--AI team performance is attributed to the fact that its evaluation is based on the proxy task of how well humans can accurately predict AI's decisions (or its decision boundaries). The proxy task situation differs from directly evaluating how well humans and AI jointly perform as a team.
They also pointed out that explanations from AI can be inductive (e.g., example-based explanations) or deductive (e.g., rule-based explanations). 
While inductive explanations are less cognitively demanding, deductive explanations are more cognitively demanding. 
In real decision-making situations, humans tend to avoid analytical thinking, which is cognitively demanding. 
In the proxy task, humans explicitly pay attention to and deliberate the AI's deductive explanations, which creates a different situation from an actual decision-making scenario.

Regarding cognitive loads on explanations from AI, Bunica et al.~\cite{bucinca2021} pointed out that people tend to create heuristics to decide whether to follow AI suggestions to avoid the cognitive loads required to understand explanations from AI.
Consequently, as discussed earlier, humans tend to unconditionally trust AI, which leads to the inferiority of human--AI team performance. 
To address this issue, they propose introducing cognitive forcing functions in human--AI collaboration, which encourages humans to engage in analytical thinking about AI's explanations.
Cognitive forcing functions are defined as ``interventions that are applied at the decision-making time to disrupt heuristic reasoning and thus cause the person to engage in analytical thinking''. They exemplify several strategies for such cognitive forcing functions, including ``asking the person to make a decision before seeing the AI's recommendations'', ``slowing down the process'', and ``letting the person choose whether and when to see the AI recommendation''.
In a different study, Green et al.~\cite{green2019principles} pointed out that even simple cognitive forcing, which allows people to make their own decisions before presenting the AI's decisions, improves decision-making by a human--AI team.
The implications of these studies indicate that introducing the cognitive forcing functions in human--AI collaboration would reduce the unconditional or excessive trust in AI caused by people avoiding their cognitive loads, and improve human--AI team performance.

\subsection{Evaluating Trust in Collaborative Human Computation}
As discussed earlier, explanations from AI can increase the interpretability and transparency of a model, 
which in turn leads to establishing trust in the model.
Explanations from AI also help people build more accurate mental models of a system, resulting in enhancing trust.
In the following section, we discuss how to evaluate trust in the context of human--AI collaboration.

Vereshak et al.~\cite{vereschak2021} surveyed how to evaluate trust in human--AI decision-making.
According to the definition of trust, ``an attitude that an agent will achieve an individual's goal in a situation characterized by uncertainty and vulnerability''~\cite{lee2004trust}, they proposed three theoretical elements of trust, ``vulnerability,'' ``positive expectations,'' and ``attitude.''
Vulnerability indicates that a person is in a situation where the outcome of a decision involves uncertainty and potentially negative or undesired consequences.
From a social-cognitive viewpoint, they claim that trust is not ``behavior'' but ``attitude'' as trust cannot be systematically replaced by behavior and cannot be fully observed by a third party.
They also pointed out that it is appropriate to discuss ``confidence'' instead of trust in situations where there are no elements of vulnerability and discuss ``distrust'' in the situations with no positive expectations.
According to their organization of the concepts, ``reliance'' refers to whether a person follows the system's suggestions, ``compliance'' refers to whether a person asks the system for suggestions. Those are not ``attitude'' but actual ``behavior.'' Then, they proposed objective, quantitative behavioral indicators for evaluating trust as shown in Table~\ref{tab:trustmeasures}.
In addition, Lai et al.~\cite{lai2021towards} proposed both subjective and objective indicators for evaluating trust and reliance in human--AI decision-making as shown in Table~\ref{tab:trustreliance}.

\begin{table}
  \caption{Trust-related Behavioral Measures~\cite{vereschak2021}}
  \label{tab:trustmeasures}
  \begin{tabular}{l|l}
     \hline
     Measures & Definition   \\
     \hline
     \hline
    Decision time & how fast a recommendation is accepted \\
        \hline
    Compliance & the number of times participants follow the systems' recommendations\\
        \hline
    Reliance & the number of times participants asked for a recommendation \\
        \hline
    Agreement/Disagreement & how quickly a recommendation is accepted\\
        \hline
    Switch ratio & the number of times a participant who initially
disagreed with the system \\
                & decided to follow its recommendation in the end\\
        \hline
\end{tabular}
\end{table}

\begin{table}
  \caption{Evaluation Metrics of Trust and Reliance~\cite{lai2021towards}}
  \label{tab:trustreliance}
  \begin{tabular}{l|l}
    \hline
    Metrics & Examples  \\
    \hline
    \hline
    Subjective & self-reported trust, model confidence/acceptance, self-reported agreement/reliance, \\
    & perceived accuracy, perceived capability/benevolence/integrity, usage intention/willingness \\
        \hline
    Objective & agreement/acceptance of model suggestions, switch, weight of advice, \\
    & model influence (difference between conditions), disagreement/deviation, choice to use the model,\\
    & over-reliance, under-reliance, appropriate reliance\\
    \hline
\end{tabular}
\end{table}

We discussed earlier that explanations would lead to humans unconditionally trusting AI and degrading the human--AI team performance. 
To address this issue, the importance of making humans think analytically about AI's explanations by introducing cognitive forcing functions has been proposed.
By paying more attention to AI' predictions and explanations, humans can carefully examine the correctness or incorrectness of AI' predictions.
This also leads to the development of accurate mental models of AI, which fosters trust in AI.
The importance of such trust calibration (knowing when AI is wrong and when to trust/distrust AI) is commonly discussed in the context of machine automation~\cite{pop2015individual}.
Zhang et al.~\cite{zhang2020effect} show that presenting confidence scores (the chance that AI is correct, e.g., the probability of every single prediction) on AI's predictions helps calibrate trust (they evaluate using behavioral indicators of switch percentage and agreement percentage) in AI.
Their experimental results showed that AI's local explanations  (feature weights-based) did not contribute to trust calibration because of the information overload that AI brings.
They also pointed out that calibrating trust alone is insufficient to improve human--AI decision-making performance and that humans should have relevant domain knowledge to appropriately complement AI errors.

Finally, regarding the relationship between trust and ethics, Flathmann et al.~\cite{flathmann2021} pointed out that ethical AI gains trust from humans and therefore improves human--AI team performance to solve problems.
Ethical AI is also important in fairness of human--AI decision-making.
They identified the ethical requirements for human--AI team: (1) AI has at least partial autonomy in decision-making; (2) AI has a clear role in the team; and (3) AI is interdependent with humans on its activities and outcomes.
They then proposed a model of ethical human--AI teamwork, in which humans share their ethical ideology and AIs share their joint team-specific ethical ideology with each other. 
Such sharing of ethical ideology allows for building more robust trust between humans and AI as teammates.
Ethical AI should gain trust over time by fulfilling its role in the team.

%% file: sec/challenges.tex
Thus far, we have summarized previous research related to trustworthy human computation from the perspective of achieving this goal.
Finally, we discuss four research directions for further development of trustworthy human computation: (1) tools and libraries; (2) secure and distributed human computation; (3) bias control; (4) reciprocal human computation; and (5) grand challenges.

\subsection{Tools and Libraries for Trustworthy Human Computation}
In Chapter 2, we survey the quality evaluation criteria for human computation from various perspectives such as reliability, availability, and serviceability. For example, methods for reliability include estimating worker ability, ground truth answers, as well model parameters. Methods for availability motivate active participation of workers. There are also techniques for serviceability such as human computation programming and workflow control. The current remarkable expansion of deep learning is supported by frameworks such as PyTorch and TensorFlow, which enable the use of deep learning without considering the details of learning algorithms and other complications. The development and standardization of these techniques as tools and libraries is an important factor for human computation to spread as a trustworthy problem-solving infrastructure. Specifically, although several research prototypes exist for workflow modeling and programming languages to improve the serviceability of human computation, none of them are widely used on commercial platforms. Similar to general software development, standardization of modeling methods and languages are important for proliferation of human computation.

\subsection{Secure and Distributed Human Computation}
In addition to RAS, security is another factor that improves the trustworthiness of human computation as computational systems. Current human computation platforms operate on centralized servers. Because requesters and workers information is concentrated on these servers, there are concerns about the leakage of such information if the operator is untrustworthy. To address such security problems, a decentralized platform using blockchain was developed~\cite{li2019crowdbc}. On this platform, human computation processes such as user registration, task registration, and task assignment are realized without a centralized server. It will be important to securely connect multiple human computation platforms to work together to achieve complex workflows.

\subsection{Control of Various Biases}
In Chapter 2, we saw that humans have an overwhelmingly larger range of biases and variances than conventional computers pertaining to various factors such as attributes, knowledge, ability, and motivation, and these factors greatly affect the results of human computation. Among these biases and variances, there is still room for further development of techniques to improve the stability of human computation by influencing the suppressing of bad biases and variances and encouraging good ones. Above all, cognitive biases are unique to humans. Although the reproducibility and universality of various types of cognitive biases are still under debate in the field of psychology, one important research direction would be to examine cognitive biases in the context of human computation and to use them for controlling human biases and variances.

In Chapter 3, we examined attempts to make human computation trustworthy in society. Because social biases are reflected in data, which directly affect the trustworthiness of human computation, the control of these biases is clearly important. On the other hand, regardless of how successful social biases can be technically removed, technology itself will not lead to social trust if it does not gain a certain level of understanding from society. Although this survey also introduced the XAI (explainability of AI), it is not yet clear how this can be connected to the accountability of fairness of human computation.

\subsection{Reciprocal Human Computation}
In Chapter 4, we examined the idea of an AI and groups of humans working together to solve difficult problems. This entails going beyond the limited/one-way trust described in Chapters 2 and 3, and building a mutually trusting relationship between AIs and humans. The construction of mutually beneficial and sustainable relationships, in which both parties can amplify their abilities and grow together through mutual trust is ideal for human computation.
One of our major challenges for the future is to find common patterns by accumulating and analyzing successes and failures in various real-world human computation applications, and to develop this into a systematic theory of system construction that transcends individual cases.

\subsection{Grand Challenges in Human Computation}
Human computation is a complex and interdisciplinary theme that develops through problems and solutions in various fields and through the fusion of knowledge. Just as the ``grand challenges'' are symbolic goals of AI, such as winning against top-tier players in board games or determining the three-dimensional structure of proteins in bioinformatics, and they have served as driving forces to propelling the development of the field, the grand challenges of human computation are also expected to play an important role in the further development of human computation. For example, research is among the most intelligent of human activities. Accelerating scientific discovery through collaboration between large numbers of AI and humans, contributing to the achievement of common human goals such as the SDGs (Sustainable Development Goals), solving pressing global issues such as COVID, and developing human resources through these efforts are major milestones which demonstrate the potential of human computation, which has previously been focused on performing relatively simple tasks.

%% file: sec/conclusion.tex
In this survey, we organized various efforts related to the trustworthiness of human computation, which is closely related to both ``human populations as users'' and ``human populations as driving forces,'' to establish mutual trust between these two human populations. First, we organized the existing research related to the trustworthiness of human computation as computing engines, that is, the trust experienced by humans of AI, by mapping them to the RAS trustworthiness measures in conventional computer systems. Thereafter, we summarized past discussions on the trustworthiness between human computation systems and users or participants, focusing on ethics such as fairness and privacy. We also summarized the discussion on human-in-the-loop human computation from the viewpoint of problem solving through collaboration between groups of humans and AI, as well as the viewpoint of hybrid intelligence, in which both parties cooperate more closely and grow together. 

We believe that human computation, which involves the cooperation of many humans to solve problems that are difficult to solve by AI alone, is a concept that will serve as a foundation for exploring the use of AI with respect to humans, and the issues and research directions for trustworthy human computation described at the end of this survey will become a key factor in realizing human-centered AI.